\documentclass[letterpaper]{article} 
\usepackage{aaai2026}  
\usepackage{times}  
\usepackage{helvet}  
\usepackage{courier}  
\usepackage[hyphens]{url}  
\usepackage{graphicx} 
\usepackage{natbib}  
\usepackage{caption} 
\usepackage{algorithm}
\usepackage{algorithmic}

\usepackage{amsmath, amssymb, amsfonts, bm}
\usepackage{graphicx}
\usepackage{subfigure}
\usepackage{caption}
\usepackage{booktabs}    
\usepackage{multirow}
\usepackage{tabularx}
\usepackage{makecell}

\usepackage{newfloat}
\usepackage{listings}
\DeclareCaptionStyle{ruled}{labelfont=normalfont,labelsep=colon,strut=off} 
\floatstyle{ruled}
\newfloat{listing}{tb}{lst}{}
\floatname{listing}{Listing}
\title{PTSM: Physiology-aware and Task-invariant Spatio-temporal Modeling for Cross-Subject EEG Decoding}

\author {
    Changhong Jing\textsuperscript{\rm 1,\rm 2},
    Yan Liu\textsuperscript{\rm 1},
    Shuqiang Wang\textsuperscript{\rm 2},
    Bruce X.B. Yu\textsuperscript{\rm 3},
    Gong Chen\textsuperscript{\rm 4},
    Zhejing Hu\textsuperscript{\rm 1},
    Zhi Zhang\textsuperscript{\rm 1},
    Yanyan Shen\textsuperscript{\rm 2}
}
\affiliations {
    \textsuperscript{\rm 1}The Hong Kong Polytechnic University\\
    \textsuperscript{\rm 2}Shenzhen Institute of Advanced Technology, Chinese Academy of Sciences\\
    \textsuperscript{\rm 3}Zhejiang University\\
    \textsuperscript{\rm 4}FireTorch Partners
}

\begin{document}

\maketitle

\begin{abstract}
Cross-subject electroencephalography (EEG) decoding remains a fundamental challenge in brain–computer interface (BCI) research due to substantial inter-subject variability and the scarcity of subject-invariant representations. This paper proposed \textbf{PTSM} (Physiology-aware and Task-invariant Spatio-temporal Modeling), a novel framework for interpretable and robust EEG decoding across unseen subjects. PTSM employs a dual-branch masking mechanism that independently learns personalized and shared spatio-temporal patterns, enabling the model to preserve individual-specific neural characteristics while extracting task-relevant, population-shared features. The masks are factorized across temporal and spatial dimensions, allowing fine-grained modulation of dynamic EEG patterns with low computational overhead. To further address representational entanglement, PTSM enforces information-theoretic constraints that decompose latent embeddings into orthogonal task-related and subject-related subspaces. The model is trained end-to-end via a multi-objective loss integrating classification, contrastive, and disentanglement objectives. Extensive experiments on cross-subject motor imagery datasets demonstrate that PTSM achieves strong zero-shot generalization, outperforming state-of-the-art baselines without subject-specific calibration. Results highlight the efficacy of disentangled neural representations for achieving both personalized and transferable decoding in non-stationary neurophysiological settings.
\end{abstract}

\section{Introduction}

Cross-subject electroencephalography (EEG) decoding \cite{yi2023learning,yan2024darnet} is a long-standing challenge in brain--computer interface (BCI) research, largely due to the high inter-subject variability in brain anatomy, electrode placement, and neural response patterns \cite{mentzelopoulosneural, melbaum2022conserved}. This variability leads to significant distribution shifts between subjects, impairing the performance of models trained on one set of individuals when applied to unseen users \cite{wang2021segregation, kronemer2022human}. As a result, building generalizable, calibration-free EEG decoding \cite{duan2023dewave,zhang2023brant,yang2023biot} systems requires representations that are both discriminative and invariant to individual differences.

Early successes in EEG decoding leveraged end-to-end deep learning models, such as EEGNet \cite{lawhern2018eegnet} and DeepConvNet \cite{schirrmeister2017deep}, which extract spatial-temporal features from raw EEG signals. These models perform well within-subject but degrade under cross-subject conditions due to their inability to handle distributional shift. To mitigate this, methods such as sparse Bayesian learning \cite{wang2023sparse} and SPD-based normalization \cite{kobler2022spd} introduce structural regularization and domain-specific priors \cite{wang2024eegpt}. Yet, these approaches often rely on assumptions about target domain access or lack interpretability.

Parallel to these developments, domain adaptation methods have emerged as a powerful strategy for improving generalization \cite{mellot2024geodesic}. Techniques such as Domain-Adversarial Neural Networks (DANN) \cite{ganin2016domain} align feature distributions via adversarial training, while spectral-temporal alignment methods \cite{liu2024boosting} further refine cross-domain invariance. However, these solutions typically assume access to unlabeled or labeled target subject data during training, which is unrealistic in many practical BCI scenarios \cite{zhang2012generalization, ben2006analysis}.

Concurrently, neuroscience-inspired frameworks attempt to model cortical dynamics using physiologically motivated structures. For example, hierarchical cortical control \cite{veniero2021top} and large-scale integration and segregation of brain networks \cite{shine2016dynamics, mohr2016integration, wang2021segregation} suggest distinct roles for domain-general and subject-specific neural pathways. Leveraging such priors, recent advances include microstate-based modeling \cite{wu2024network}, sparse neuronal representations \cite{yoshida2020natural,shen2024robust}, and structural signal disentanglement \cite{gokcen2022disentangling,gnassounou2023convolution}. These methods bring interpretability but are often limited in scalability and representational flexibility.

More recently, representation disentanglement approaches have gained traction. CSLP-AE \cite{norskov2023cslp} uses contrastive latent permutation autoencoding to separate subject-related and task-relevant features. Similarly, mask-guided neural modeling \cite{zheng2024discrete} and mutual information-based disentanglement \cite{dunion2023conditional} promote modular representation learning, offering improved interpretability and generalization \cite{yuan2023ppi}. However, they still leave unresolved challenges in robustly capturing both individualized and shared neural dynamics.

Nonetheless, despite considerable progress across various approaches, fundamental challenges remain in achieving truly generalizable EEG decoding \cite{wang2023contrast,ma2024learning,tu2024dmnet}. Motivated by these persistent challenges, we proposed \textbf{PTSM}, a unified framework for cross-subject EEG decoding. PTSM introduces a dual-branch modulation mechanism that explicitly factorizes EEG signals into personalized and task-general pathways. It incorporates hierarchical spatio-temporal attention modules inspired by cortical processing theories \cite{melbaum2022conserved,veniero2021top, shine2016dynamics}, and employs feature disentanglement guided by soft orthogonality and mutual information regularization. PTSM is optimized via a multi-objective loss that integrates classification, contrastive learning, and structural sparsity, leading to superior cross-subject generalization without requiring target subject data.

\noindent\textbf{Contributions.} This work introduces a unified framework for cross-subject EEG decoding with the following core contributions:

\begin{itemize}
\item \textbf{Spatio-Temporal Alignment and Personalization.}
A dual-path architecture is proposed to separately model subject-specific and task-invariant EEG dynamics. The individualized branch learns temporal and spatial relevance patterns specific to each subject. The shared branch identifies common neural structures across subjects related to the task. A learnable fusion strategy integrates both sources of information. Contrastive regularization is introduced to ensure representational complementarity. Factorization into temporal and spatial modules reduces redundancy and enhances interpretability.

\item \textbf{Task-Specific and Subject-Specific Feature Decoupling.}
A feature decomposition strategy is developed to isolate task-discriminative and subject-characteristic components. Two latent subspaces are constructed for representing cognitive and identity-related factors. Disentanglement is encouraged using soft regularization techniques. These include vector-level orthogonality covariance-level decorrelation, mutual information maximization, and sparsity regularization. This design improves subject-invariant representation quality while preserving personalized expressiveness.

\item \textbf{Unified Multi-Objective Optimization.}
A comprehensive optimization framework is formulated to balance classification, disentanglement, and contrastive supervision. Cross-entropy losses guide task and subject label prediction. Contrastive objectives promote instance-level separation in latent spaces. Additional constraints on activation sparsity and magnitude regulate the structure of learned spatio-temporal filters. The integration of these objectives supports the learning of modular, transferable, and generalizable EEG representations.
\end{itemize}

\section{Method}

\subsection{Problem Formulation}

Let $\mathcal{D} = {(\mathbf{x}i, y_i, s_i)}{i=1}^N$ be a multichannel EEG dataset, where $\mathbf{x}_i \in \mathbb{R}^{C \times T}$ is a trial with $C$ channels and $T$ time points, $y_i \in {1, \dots, K}$ denotes the task label, and $s_i \in {1, \dots, S}$ indicates the subject identity. The goal is to learn a prediction function $f: \mathbb{R}^{C \times T} \to {1, \dots, K}$ that generalizes to unseen subjects. A key challenge lies in the high inter-subject variability in EEG signals. To address this, we aim to disentangle each trial’s latent representation into orthogonal task-related and subject-related components, enabling robust cross-subject decoding.

\subsection{Framework Overview}

\begin{figure*}[h]
 \centering
 \includegraphics[width=0.8\linewidth]{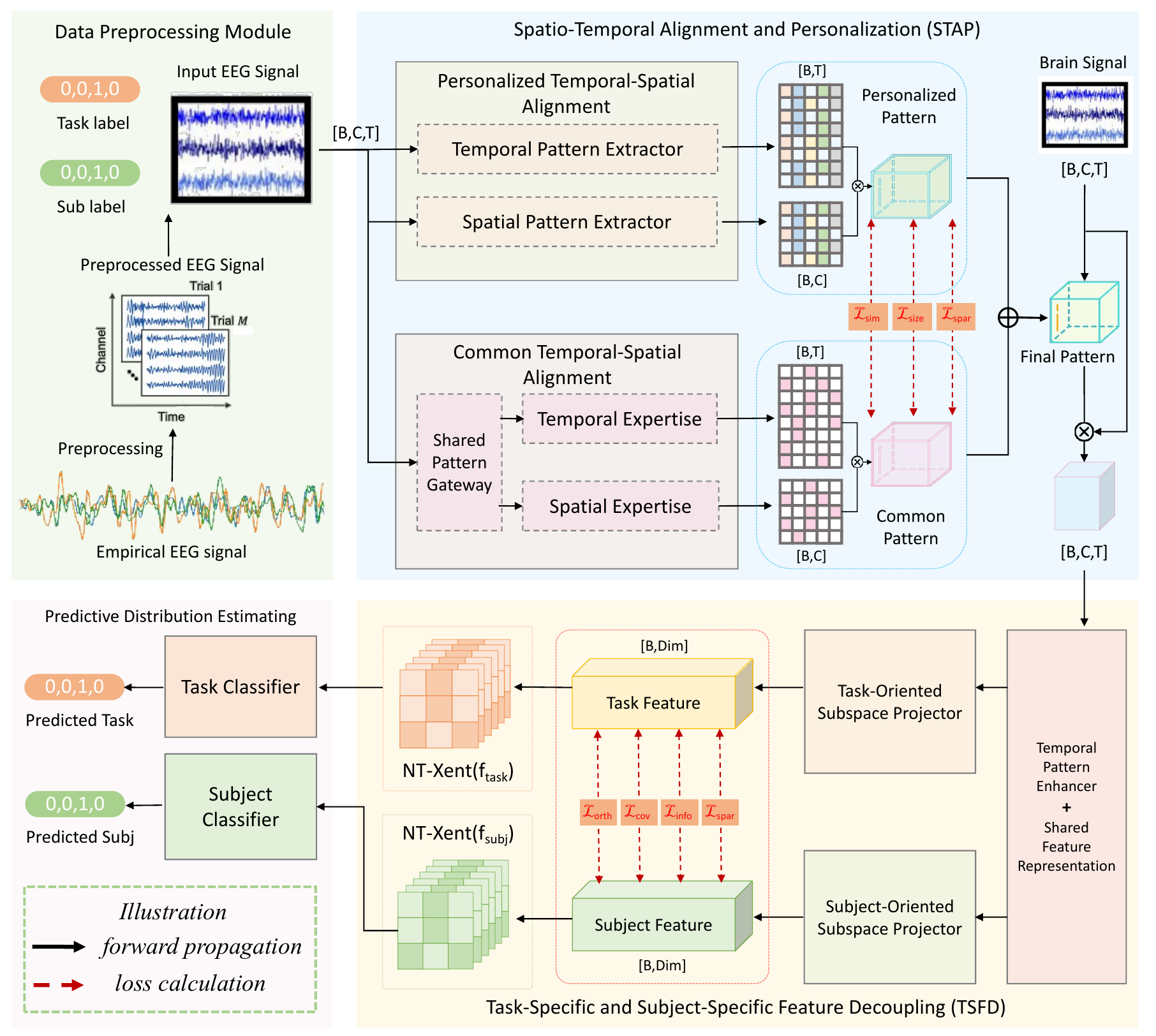}
 \caption{
 \textbf{Overview of the PTSM Framework.}
 The proposed model consists of two core modules:
 (1) a Spatio-Temporal Alignment and Personalization (STAP) module that adaptively captures personalized and shared EEG patterns across time and space, and
 (2) a Task-Specific and Subject-Specific Feature Decoupling (TSFD) module that disentangles task-related and subject-related components via multi-view subspace projections.
 Black arrows indicate forward propagation, and red dashed arrows indicate loss calculations.
 }
 \label{fig:ptsm_framework}
 \end{figure*}

The PTSM framework integrates personalized and transferable feature modeling via two core modules: Spatio-Temporal Alignment and Personalization (STAP), and Task-Specific and Subject-Specific Feature Decoupling (TSFD). Figure~\ref{fig:ptsm_framework} provides a visual overview of the overall design.

\subsection{Spatio-Temporal Alignment and Personalization (STAP)}

The STAP module extracts both individualized and task-invariant neural patterns by learning separate temporal and spatial relevance maps. A dual-branch structure produces personalized and shared attention weights, which are adaptively fused to form final spatio-temporal modulation masks. Each EEG input $\mathbf{x} \in \mathbb{R}^{C \times T}$ is first modulated by both spatial and temporal masks before feature encoding.

\paragraph{Personalized Pattern.} The personalized mask captures individual-specific dynamics. It is factorized into a spatial mask $\mathbf{M}_s^p \in [0,1]^C$ and a temporal mask $\mathbf{M}_t^p \in [0,1]^T$, each learned via neural functions:
\begin{align}
    \mathbf{M}_s^p &= \sigma(g_s^p(\mathbf{x})) \\
    \mathbf{M}_t^p &= \sigma(g_t^p(\mathbf{x}))
\end{align}
where $g_s^p(\cdot)$ and $g_t^p(\cdot)$ are MLP and ConvNet-based modules, and $\sigma(\cdot)$ is the sigmoid activation to constrain values to $[0,1]$.

\paragraph{Common Pattern.} To model population-shared representations, a separate branch learns spatial and temporal masks that reflect task-relevant but subject-invariant patterns:
\begin{align}
    \mathbf{M}_s^c &= \sigma(g_s^c(\mathbf{x})) \\
    \mathbf{M}_t^c &= \sigma(g_t^c(\mathbf{x}))
\end{align}

\paragraph{Pattern Fusion.} The final masks are obtained via a convex combination of personalized and common maskss:
\begin{align}
    \mathbf{M}_s &= \beta \cdot \mathbf{M}_s^p + (1 - \beta) \cdot \mathbf{M}_s^c \\
    \mathbf{M}_t &= \alpha \cdot \mathbf{M}_t^p + (1 - \alpha) \cdot \mathbf{M}_t^c
\end{align}
where $\alpha, \beta \in [0,1]$ are scalar weights. The masked EEG is then computed as:
\begin{equation}
    \mathbf{x}^{\text{masked}} = \mathbf{x} \odot \mathbf{M}_s \odot \mathbf{M}_t
\end{equation}
which enhances subject-relevant and task-consistent information in the input.

\subsection{Mask Contrastive Learning}

To explicitly encourage the personalized and common masks to capture complementary information, we incorporate a contrastive objective that penalizes their similarity. Given temporal and spatial masks $\mathbf{M}^p = \mathbf{M}_t^p \otimes \mathbf{M}_s^p$ and $\mathbf{M}^c = \mathbf{M}_t^c \otimes \mathbf{M}_s^c$, we define a cosine similarity-based loss:
\begin{equation}
	\mathcal{L}_{\text{sim}} = \frac{1}{N} \sum_{i=1}^{N} \frac{\mathbf{M}_i^p \cdot \mathbf{M}_i^c}{\|\mathbf{M}_i^p\| \cdot \|\mathbf{M}_i^c\|}
\end{equation}

Additionally, sparsity is encouraged to make masks more interpretable and focus attention on localized regions:
\begin{equation}
	\mathcal{L}_{\text{sparse}} = \frac{1}{N} \sum_{i=1}^{N} \left( \|\mathbf{M}_i^p\|_1 + \|\mathbf{M}_i^c\|_1 \right)
\end{equation}

To control the effective size of the masks and avoid trivial solutions (e.g., all-one or all-zero masks), we introduce a size regularization term that constrains the mean mask activation to a predefined target $\alpha$:
\begin{equation}
	\mathcal{L}_{\text{size}} = \frac{1}{N} \sum_{i=1}^{N} \left( \left| \frac{1}{D} \sum_{d=1}^{D} \mathbf{M}_{i,d}^p - \alpha \right| + \left| \frac{1}{D} \sum_{d=1}^{D} \mathbf{M}_{i,d}^c - \alpha \right| \right)
\end{equation}

where $D = C \times T$ is the total number of spatiotemporal positions.

This formulation explicitly regularizes the learning of interpretable, diverse, and complementary mask structures, which are critical for capturing both personalized and generalizable EEG patterns.

\subsection{Task-Specific and Subject-Specific Feature Decoupling (TSFD)}

The masked EEG signal $\mathbf{x}^{\text{masked}} \in \mathbb{R}^{C \times T}$ is processed by a temporal encoder and further decomposed into orthogonal subspaces that separately capture task-relevant and subject-specific information.

\paragraph{Temporal Feature Extraction.}
We apply a temporal encoder $F_{\text{temp}}$ composed of 1D convolutional layers followed by batch normalization, ELU activation, dropout, and adaptive pooling to capture local temporal dynamics:
\begin{equation}
    \mathbf{h}_{\text{temp}} = F_{\text{temp}}(\mathbf{x}^{\text{masked}})
\end{equation}

\paragraph{Shared Representation Encoder.}
The extracted temporal features are further mapped into a latent shared space via a fully connected encoder:
\begin{equation}
    \mathbf{h}_{\text{shared}} = F_{\text{shared}}(\mathbf{h}_{\text{temp}})
\end{equation}
where $F_{\text{shared}}$ is implemented as an MLP with nonlinear activation.

\paragraph{Latent Space Disentanglement.}
To separate the latent representation into distinct task and subject components, $\mathbf{h}_{\text{shared}}$ is projected into two independent subspaces:
\begin{align}
    \mathbf{f}^{\text{task}} &= F_{\text{task}}(\mathbf{h}_{\text{shared}}) \\
    \mathbf{f}^{\text{subj}} &= F_{\text{subj}}(\mathbf{h}_{\text{shared}})
\end{align}
where $F_{\text{task}}$ and $F_{\text{subj}}$ are lightweight projection heads that produce task-discriminative and subject-discriminative features, respectively.

\subsection{Orthogonal Representation Constraints}

To encourage statistical independence and disentanglement between the task-related and subject-related feature spaces, we impose four complementary constraints: vector-level orthogonality, covariance decorrelation, mutual information maximization, and sparsity regularization.

\paragraph{Vector-level Orthogonality.}
We explicitly enforce orthogonality between the task and subject features at the instance level by minimizing their cosine similarity:
\begin{equation}
    \mathcal{L}_{\text{orth}} = \frac{1}{N} \sum_{i=1}^{N} \frac{|\mathbf{f}_i^{\text{task}} \cdot \mathbf{f}_i^{\text{subj}}|}{\|\mathbf{f}_i^{\text{task}}\|_2 \cdot \|\mathbf{f}_i^{\text{subj}}\|_2}
\end{equation}

\paragraph{Covariance-level Decorrelation.}
To decorrelate the task and subject representations at the population level, we minimize the Frobenius norm of their covariance matrix:
\begin{equation}
    \mathcal{L}_{\text{cov}} = \frac{\|\text{Cov}(\mathbf{f}^{\text{task}}, \mathbf{f}^{\text{subj}})\|_F}{\|\text{Cov}(\mathbf{f}^{\text{task}})\|_F \cdot \|\text{Cov}(\mathbf{f}^{\text{subj}})\|_F}
\end{equation}
where $\text{Cov}(\cdot)$ denotes the empirical covariance matrix over a mini-batch and $\|\cdot\|_F$ is the Frobenius norm.

\paragraph{Mutual Information Maximization.}
To preserve the information content in both feature spaces, we encourage variance retention by maximizing the trace of their respective covariance matrices:
\begin{equation}
    \mathcal{L}_{\text{info}} = -\left(\text{tr}(\text{Cov}(\mathbf{f}^{\text{task}})) + \text{tr}(\text{Cov}(\mathbf{f}^{\text{subj}}))\right)
\end{equation}
where $\text{tr}(\cdot)$ is the trace operator.

\paragraph{Sparsity Regularization.}
To enhance interpretability and promote compact representations, we impose $\ell_1$ regularization on both feature vectors:
\begin{equation}
    \mathcal{L}_{\text{sparse}}^{\prime} = \sum_{i=1}^{N} \left(\|\mathbf{f}_i^{\text{task}}\|_1 + \|\mathbf{f}_i^{\text{subj}}\|_1\right)
\end{equation}

\subsection{Unified Multi-objective Optimization}

To jointly optimize the objectives of task decoding, subject separation, representation disentanglement, and mask complementarity, we adopt a unified multi-loss formulation. The total training loss is defined as:


\begin{align}
    \mathcal{L}_{\text{total}} = \mathcal{L}_{\text{task}} 
    + \lambda_{\text{subj}} \mathcal{L}_{\text{subj}} 
    + \lambda_{\text{decouple}} \mathcal{L}_{\text{decouple}} \nonumber \\
    + \lambda_{\text{mask}} \mathcal{L}_{\text{mask}} 
    + \lambda_{\text{contrast}} \mathcal{L}_{\text{contrast}}
\end{align}


\paragraph{Classification objectives.} The model employs cross-entropy losses for both task and subject prediction. These losses drive the learning of discriminative representations and provide complementary supervision for the decoupled subspaces.

\paragraph{Representation disentanglement.} The disentanglement loss includes vector orthogonality, batch-level covariance decorrelation, mutual information maximization, and sparsity regularization:

\begin{align}
    \mathcal{L}_{\text{decouple}} = \lambda_{\text{orth}} \mathcal{L}_{\text{orth}} 
    + \lambda_{\text{cov}} \mathcal{L}_{\text{cov}} \nonumber \\
    + \lambda_{\text{info}} \mathcal{L}_{\text{info}} 
    + \lambda_{\text{sparse}}^{\prime} \mathcal{L}_{\text{sparse}}^{\prime}
\end{align}

\paragraph{Contrastive supervision.} The objective $\mathcal{L}_{\text{contrast}}$ maximizes discriminability among different samples in latent space using normalized temperature-scaled cross entropy. 

\paragraph{Mask regularization.} The mask-specific loss $\mathcal{L}_{\text{mask}}$ encourages diversity, sparsity, and interpretability of spatial and temporal attention maps.

The final mask contrastive loss combines these objectives:
\begin{equation}
	\mathcal{L}_{\text{mask}} = \lambda_{\text{sim}} \mathcal{L}_{\text{sim}} + \lambda_{\text{sparse}} \mathcal{L}_{\text{sparse}} + \lambda_{\text{size}} \mathcal{L}_{\text{size}}
\end{equation}
with $\lambda_{\text{sim}}, \lambda_{\text{sparse}}, \lambda_{\text{size}}$ controlling the relative importance of each component.

\medskip

All $\lambda$ terms are hyperparameters that control the trade-off between the objectives and are tuned via grid search on a validation set. Algorithm~\ref{alg:ptsm} summarizes the training procedure of PTSM. 


\begin{algorithm}[t]
\caption{\textsc{PTSM Training Algorithm}}
\label{alg:ptsm}
\begin{algorithmic}[1]
\REQUIRE Mini-batch $\mathcal{B} = \{(\mathbf{x}_i, y_i, s_i)\}_{i=1}^N$
\ENSURE Updated parameters $\theta$
\FORALL{$(\mathbf{x}_i, y_i, s_i) \in \mathcal{B}$}
    \STATE $\mathbf{M}_t^p \leftarrow \sigma(g_t^p(\mathbf{x}_i))$ 
    \STATE $\mathbf{M}_s^p \leftarrow \sigma(g_s^p(\mathbf{x}_i))$ 
    \STATE $\mathbf{M}_t^c \leftarrow \sigma(g_t^c(\mathbf{x}_i))$ 
    \STATE $\mathbf{M}_s^c \leftarrow \sigma(g_s^c(\mathbf{x}_i))$ 
    \STATE $\mathbf{M}_t \leftarrow \alpha \mathbf{M}_t^p + (1{-}\alpha) \mathbf{M}_t^c$ 
    \STATE $\mathbf{M}_s \leftarrow \beta \mathbf{M}_s^p + (1{-}\beta) \mathbf{M}_s^c$ 
    \STATE $\mathbf{x}^{\text{masked}} \leftarrow \mathbf{x}_i \odot \mathbf{M}_t \odot \mathbf{M}_s$ 
    \STATE $\mathbf{h}_{\text{temp}} \leftarrow F_{\text{temp}}(\mathbf{x}^{\text{masked}})$ 
    \STATE $\mathbf{h}_{\text{shared}} \leftarrow F_{\text{shared}}(\mathbf{h}_{\text{temp}})$
    \STATE $\mathbf{f}_i^{\text{task}} \leftarrow F_{\text{task}}(\mathbf{h}_{\text{shared}})$
    \STATE $\mathbf{f}_i^{\text{subj}} \leftarrow F_{\text{subj}}(\mathbf{h}_{\text{shared}})$ 
    \STATE $\mathbf{p}_i^{\text{task}} \leftarrow C_{\text{task}}(\mathbf{f}_i^{\text{task}})$ 
    \STATE $\mathbf{p}_i^{\text{subj}} \leftarrow C_{\text{subj}}(\mathbf{f}_i^{\text{subj}})$ 
\ENDFOR
\STATE $\mathcal{L}_{\text{task}} \leftarrow \text{CE}(\{\mathbf{p}_i^{\text{task}}, y_i\})$ 
\STATE $\mathcal{L}_{\text{subj}} \leftarrow \text{CE}(\{\mathbf{p}_i^{\text{subj}}, s_i\})$
\STATE $\mathcal{L}_{\text{decouple}} \leftarrow \lambda_{\text{orth}} \mathcal{L}_{\text{orth}} + \lambda_{\text{cov}} \mathcal{L}_{\text{cov}} + \lambda_{\text{info}} \mathcal{L}_{\text{info}} + \lambda_{\text{sparse}} \mathcal{L}_{\text{sparse}}$ 
\STATE $\mathcal{L}_{\text{mask}} \leftarrow \lambda_{\text{sim}} \mathcal{L}_{\text{sim}} + \lambda_{\text{sparse}} \mathcal{L}_{\text{sparse}} + \lambda_{\text{size}} \mathcal{L}_{\text{size}}$ 
\STATE $\mathcal{L}_{\text{contrast}} \leftarrow \lambda_{\text{task}} \mathcal{L}_{\text{contrast-task}} + \lambda_{\text{subj}} \mathcal{L}_{\text{contrast-subj}}$ 
\STATE $\mathcal{L}_{\text{total}} \leftarrow \mathcal{L}_{\text{task}} + \lambda_{\text{subj}} \mathcal{L}_{\text{subj}} + \lambda_{\text{decouple}} \mathcal{L}_{\text{decouple}} + \lambda_{\text{mask}} \mathcal{L}_{\text{mask}} + \lambda_{\text{contrast}} \mathcal{L}_{\text{contrast}}$ 
\STATE $\theta \leftarrow \theta - \eta \nabla_{\theta} \mathcal{L}_{\text{total}}$
\RETURN $\theta$
\end{algorithmic}
\end{algorithm}

\section{Experiments}

\subsection{Experimental Setup}

In this section, we present the comprehensive experimental evaluation of the proposed PTSM model. The experiments are conducted on three widely used benchmark datasets, including OpenBMI (MI and ERP) \cite{lee2019eeg}, PhysioNet Motor Imagery (MI) \cite{schalk2004bci2000,goldberger2000physiobank}, and KUL Auditory Attention Detection (AAD) \cite{biesmans2016auditory}. Our primary goal is to demonstrate the performance of PTSM in cross-subject tasks, which is a critical challenge in EEG brain decoding. Additionally, we aim to provide insights into the effectiveness of our proposed dual-masking mechanism and feature decoupling strategy through extensive ablation studies. The experiments are performed under a consistent cross-subject protocol, where models are trained on data from multiple subjects and tested on unseen subjects. This setup accurately reflects real-world scenarios, ensuring the generalizability of the proposed model.

\subsection{Comparison with Existing Methods}

 \begin{table*}[h!]
 \centering
 \begin{tabular}{lcccc}
 \toprule
 \multirow{2}{*}{Method} & \multicolumn{2}{c}{Session 1} & \multicolumn{2}{c}{Session 2} \\
 \cmidrule(lr){2-3} \cmidrule(lr){4-5}
  & Pre-ACC (\%) & Post-ACC (\%) & Pre-ACC (\%) & Post-ACC (\%) \\
 \midrule
 EEGNet & 62.57 $\pm$ 0.49 & 65.39 $\pm$ 0.56 & 63.26 $\pm$ 0.56 & 66.46 $\pm$ 0.65 \\
 DeepConvNet & 64.37 $\pm$ 0.61 & 67.42 $\pm$ 0.63 & 64.94 $\pm$ 0.66 & 67.62 $\pm$ 0.51 \\
 DANN & 65.46 $\pm$ 0.43 & 68.13 $\pm$ 0.51 & 66.07 $\pm$ 0.63 & 68.68 $\pm$ 0.52 \\
 ACON & 66.51 $\pm$ 0.35 & 68.63 $\pm$ 0.43 & 67.01 $\pm$ 0.51 & 68.90 $\pm$ 0.47 \\
 CSLPAE & 67.24 $\pm$ 0.44 & 68.85 $\pm$ 0.49 & 67.59 $\pm$ 0.45 & 68.41 $\pm$ 0.44 \\
 TSMNet  & 67.83 $\pm$ 0.35 & 69.26 $\pm$ 0.38 & 68.23 $\pm$ 0.36 & 69.38 $\pm$ 0.42 \\
 SBLEST  & 68.56 $\pm$ 0.32 & 69.44 $\pm$ 0.35 & 68.71 $\pm$ 0.37 & 69.51 $\pm$ 0.38 \\
 \textbf{PTSM (Ours)} & \textbf{71.87 $\pm$ 0.49} & \textbf{72.53 $\pm$ 0.54} & \textbf{72.67 $\pm$ 0.64} & \textbf{73.26 $\pm$ 0.59} \\
 \bottomrule
 \end{tabular}
 \caption{Comparison Performance with Existing Methods}
 \label{tab:cross_session}
 \end{table*}

We begin by comparing the performance of PTSM with several existing methods, including EEGNet \cite{lawhern2018eegnet}, DeepConvNet \cite{schirrmeister2017deep}, DANN \cite{ganin2016domain}, ACON \cite{liu2024boosting}, CSLPAE \cite{norskov2023cslp}, TSMNet \cite{kobler2022spd}, and SBLEST \cite{wang2023sparse}. Table \ref{tab:cross_session} presents the results on the OpenBMI Motor Imagery (MI) task under cross-subject classification settings. The performance of PTSM can be attributed to its STAP mechanism, which effectively captures both common and personal spatio-temporal patterns, enhancing model generalization. Moreover, the TSFD strategy further improves robustness by separating subject-specific and task-related features.

\subsection{Ablation Study on STAP Mechanism}

 \begin{table*}[h!]
 \centering
 \begin{tabular}{lcccc}
 \toprule
 \multirow{2}{*}{Configuration} & \multicolumn{2}{c}{Session 1} & \multicolumn{2}{c}{Session 2} \\
 \cmidrule(lr){2-3} \cmidrule(lr){4-5}
 & Pre-ACC (\%) & Post-ACC (\%) & Pre-ACC (\%) & Post-ACC (\%) \\
 \midrule
 w/o STAP & 67.54 $\pm$ 0.45 & 67.75 $\pm$ 0.52 & 67.17 $\pm$ 0.49 & 67.84 $\pm$ 0.52 \\
 w/o PP & 68.19 $\pm$ 0.51 & 68.27 $\pm$ 0.54 & 68.93 $\pm$ 0.53 & 69.21 $\pm$ 0.49 \\
 w/o CP & 69.33 $\pm$ 0.49 & 69.46 $\pm$ 0.48 & 69.72 $\pm$ 0.56 & 69.94 $\pm$ 0.51 \\
 \textbf{PTSM (Ours)} & \textbf{71.87 $\pm$ 0.49} & \textbf{72.53 $\pm$ 0.54} & \textbf{72.67 $\pm$ 0.64} & \textbf{73.26 $\pm$ 0.59} \\
 \bottomrule
 \end{tabular}
 \caption{Ablation Study on STAP Mechanism}
 \label{tab:mask_ablation}
 \end{table*}


Table~\ref{tab:mask_ablation} summarizes the ablation study results for the STAP mechanism under different configurations. Four configurations are compared: removing the STAP module (w/o STAP), removing the personal pattern (w/o PP), removing the common pattern (w/o CP), and using the complete PTSM model. Performance is evaluated on two sessions, with results reported for both pre- and post-adaptation accuracy. The complete PTSM model achieves the highest accuracy across all sessions and conditions. Compared with the other configurations, PTSM consistently yields higher pre- and post-adaptation performance. Removing either the common or personal pattern mask leads to a clear reduction in accuracy. This finding highlights the necessity of integrating both common and personal patterns for optimal decoding performance. The ablation results demonstrate that the STAP mechanism captures task-relevant global features and individual-specific information, both of which are essential for robust EEG decoding.

\subsection{Ablation Study on TSFD Strategy}


We evaluate the effectiveness of the feature decoupling strategy through ablation experiments on key components. These components include vector orthogonality (VO), covariance independence (CO), mutual information minimization (MIM), and sparsity regularization (SR). Table~\ref{tab:decoupling_ablation} shows the results of these experiments. The results show that each decoupling constraint helps to improve model performance. The full PTSM model consistently achieves the best scores. 


Removing any of these constraints causes the model performance to decline. The most significant decrease occurs when either vector orthogonality or covariance independence is removed. These results suggest that maintaining feature independence is essential for robust cross-subject classification.

 \begin{table*}[h!]
 \centering
 \begin{tabular}{lcccc}
 \toprule
 \multirow{2}{*}{Configuration} & \multicolumn{2}{c}{Session 1} & \multicolumn{2}{c}{Session 2} \\
 \cmidrule(lr){2-3} \cmidrule(lr){4-5}
 & Pre-ACC (\%) & Post-ACC (\%) & Pre-ACC (\%) & Post-ACC (\%) \\
 \midrule
 w/o VO & 68.27 $\pm$ 0.47 & 69.33 $\pm$ 0.41 & 69.45 $\pm$ 0.54 & 70.91 $\pm$ 0.59 \\
 w/o CO & 67.84 $\pm$ 0.38 & 68.92 $\pm$ 0.63  & 68.43 $\pm$ 0.47 & 69.59 $\pm$ 0.46 \\
 w/o MIM & 68.03 $\pm$ 0.57   & 69.26 $\pm$ 0.52 & 68.75 $\pm$ 0.64 & 69.43 $\pm$ 0.55 \\
 w/o SR & 69.35 $\pm$ 0.56 & 70.57 $\pm$ 0.59  & 69.96 $\pm$ 0.61 & 70.28 $\pm$ 0.55 \\
 \textbf{PTSM (Ours)} & \textbf{71.87 $\pm$ 0.49} & \textbf{72.53 $\pm$ 0.54} & \textbf{72.67 $\pm$ 0.64} & \textbf{73.26 $\pm$ 0.59} \\
 \bottomrule
 \end{tabular}
 \caption{Ablation Study on TSFD Strategy}
 \label{tab:decoupling_ablation}
 \end{table*}

\subsection{Performance on Diverse Datasets and Tasks}

To further validate the generalization of the proposed PTSM model, we evaluate its performance on three datasets: PhysioNet Motor Imagery, KUL Auditory Attention Detection (AAD), and OpenBMI Event-Related Potential (ERP). Tables \ref{tab:physionet}, \ref{tab:kul_aad} and \ref{tab:openbmi_erp} present the cross-subject classification results on these datasets, respectively. Our model demonstrates superior accuracy, F1 score, sensitivity, and specificity across all datasets, significantly outperforming state-of-the-art methods.

The consistent performance of PTSM on these diverse datasets highlights its robustness and adaptability. Specifically, the model achieves the highest accuracy on the OpenBMI ERP dataset, reflecting its ability to generalize across different EEG tasks. This strong generalization can be attributed to the combination of dual-masking and feature decoupling, which effectively captures both global task-related patterns and individual-specific characteristics using the same training and testing configuration, with models trained on data from multiple subjects and tested on unseen subjects. The results demonstrate that PTSM consistently outperforms existing methods across all metrics, with significant improvements in both accuracy and F1 score.

 \begin{table*}[h!]
 \centering 
 \begin{tabular}{lcccc}
 \toprule
 Method & ACC (\%) & F1 (\%) & SEN (\%) & SPE (\%) \\
 \midrule
 EEGNet        & 68.43 $\pm$ 0.47 & 67.84 $\pm$ 0.46 & 69.03 $\pm$ 0.53 & 65.26 $\pm$ 0.49 \\
 DeepConvNet   & 70.35 $\pm$ 0.49 & 69.93 $\pm$ 0.48 & 70.81 $\pm$ 0.51 & 67.91 $\pm$ 0.52 \\
 DANN          & 71.54 $\pm$ 0.51 & 71.07 $\pm$ 0.45 & 72.05 $\pm$ 0.54 & 69.33 $\pm$ 0.47 \\
 ACON          & 74.27 $\pm$ 0.56 & 73.86 $\pm$ 0.49 & 74.56 $\pm$ 0.55 & 71.15 $\pm$ 0.48 \\
 CSLPAE        & 75.16 $\pm$ 0.52 & 74.62 $\pm$ 0.47 & 75.59 $\pm$ 0.53 & 72.52 $\pm$ 0.58 \\
 TSMNet        & 74.51 $\pm$ 0.48 & 74.14 $\pm$ 0.45 & 75.06 $\pm$ 0.51 & 71.37 $\pm$ 0.49 \\
 SBLEST        & 76.73 $\pm$ 0.46 & 76.27 $\pm$ 0.47 & 77.12 $\pm$ 0.48 & 73.83 $\pm$ 0.51 \\
 \textbf{PTSM (Ours)} & \textbf{79.26 $\pm$ 0.42} & \textbf{79.07 $\pm$ 0.44} & \textbf{79.54 $\pm$ 0.45} & \textbf{78.63 $\pm$ 0.43} \\
 \bottomrule
 \end{tabular}
 \caption{PTSM Performance Comparison on PhysioNet}
 \label{tab:physionet}
 \end{table*}

 \begin{table*}[h!]
 \centering
 \begin{tabular}{lcccc}
 \toprule
 Method & ACC (\%) & F1 (\%) & SEN (\%) & SPE (\%) \\
 \midrule
 EEGNet        & 54.89 $\pm$ 0.55 & 54.25 $\pm$ 0.47 & 55.07 $\pm$ 0.49 & 53.34 $\pm$ 0.63 \\
 DeepConvNet   & 60.43 $\pm$ 0.59 & 59.86 $\pm$ 0.51 & 61.05 $\pm$ 0.54 & 58.72 $\pm$ 0.52 \\
 DANN          & 62.16 $\pm$ 0.47 & 61.54 $\pm$ 0.49 & 62.87 $\pm$ 0.52 & 60.95 $\pm$ 0.51 \\
 ACON          & 64.38 $\pm$ 0.48 & 63.77 $\pm$ 0.57 & 65.06 $\pm$ 0.47 & 62.53 $\pm$ 0.49 \\
 CSLPAE        & 65.82 $\pm$ 0.51 & 65.36 $\pm$ 0.48 & 66.25 $\pm$ 0.52 & 64.03 $\pm$ 0.53 \\
 TSMNet        & 65.15 $\pm$ 0.49 & 64.74 $\pm$ 0.51 & 65.64 $\pm$ 0.55 & 63.55 $\pm$ 0.47 \\
 SBLEST        & 67.08 $\pm$ 0.47 & 66.58 $\pm$ 0.49 & 67.56 $\pm$ 0.48 & 65.86 $\pm$ 0.46 \\
 \textbf{PTSM (Ours)} & \textbf{71.54 $\pm$ 0.51} & \textbf{71.07 $\pm$ 0.54} & \textbf{72.05 $\pm$ 0.47} & \textbf{70.87 $\pm$ 0.63} \\
 \bottomrule
 \end{tabular}
 \caption{PTSM Performance Comparison on KUL AAD}
 \label{tab:kul_aad}
 \end{table*}

 \begin{table*}[h!]
 \centering
 \begin{tabular}{lcccc}
 \toprule
 Method & ACC (\%) & F1 (\%) & SEN (\%) & SPE (\%) \\
 \midrule
 EEGNet        & 67.24 $\pm$ 0.56 & 66.86 $\pm$ 0.47 & 67.52 $\pm$ 0.53 & 64.91 $\pm$ 0.46 \\
 DeepConvNet   & 69.03 $\pm$ 0.41 & 68.53 $\pm$ 0.38 & 69.53 $\pm$ 0.52 & 66.84 $\pm$ 0.55 \\
 DANN          & 70.56 $\pm$ 0.53 & 70.04 $\pm$ 0.57 & 71.25 $\pm$ 0.54 & 68.37 $\pm$ 0.54 \\
 ACON          & 73.02 $\pm$ 0.41 & 72.74 $\pm$ 0.56 & 73.53 $\pm$ 0.61 & 70.54 $\pm$ 0.59 \\
 CSLPAE        & 74.32 $\pm$ 0.42 & 74.07 $\pm$ 0.49 & 74.81 $\pm$ 0.41 & 71.77 $\pm$ 0.45 \\
 TSMNet        & 73.85 $\pm$ 0.49 & 73.36 $\pm$ 0.58 & 74.33 $\pm$ 0.55 & 71.05 $\pm$ 0.53 \\
 SBLEST        & 75.71 $\pm$ 0.54 & 75.23 $\pm$ 0.53 & 76.18 $\pm$ 0.56 & 73.03 $\pm$ 0.47 \\
 \textbf{PTSM (Ours)} & \textbf{79.06 $\pm$ 0.47} & \textbf{78.74 $\pm$ 0.48} & \textbf{79.53 $\pm$ 0.52} & \textbf{78.25 $\pm$ 0.46} \\
 \bottomrule
 \end{tabular}
 \caption{PTSM Performance Comparison on OpenBMI ERP}
 \label{tab:openbmi_erp}
 \end{table*}

\section{Discussion}


In this study, we propose \textbf{PTSM}, a novel framework for cross-subject EEG decoding. The method addresses the core challenge of inter-subject variability by disentangling subject-specific and task-relevant features, and by incorporating neuro-inspired mechanisms to enhance both personalization and generalization. 


The architecture of PTSM is inspired by conserved neurophysiological structures that are observed across individuals. Recent studies indicate that cross-subject decoding is possible because neural activity patterns in the sensorimotor cortex are preserved, even across different species and behavioral contexts~\cite{melbaum2022conserved}. 

Extensive experiments on multi-session EEG datasets demonstrate that PTSM consistently outperforms both classic deep learning models and recent baselines in terms of accuracy and F1-score. Performance gains are observed across both pre- and post-familiarization conditions, suggesting that PTSM captures temporally stable and transferable cognitive representations. Ablation studies further validate the necessity of each proposed component, particularly the decoupling constraints and dual masking mechanisms.

The model's modular structure allows direct analysis of which temporal and spatial components are emphasized per subject, offering insights into individual differences in cognitive processing. The learned masks and decoupled representations may serve as biomarkers for inter-individual variability in neural dynamics, with potential implications for personalized BCI applications and clinical neuroassessment.

Despite its empirical and theoretical strengths, PTSM has several limitations that warrant further investigation. First, the dual-path modulation mechanism increases model complexity and inference cost, which may hinder deployment in real-time systems. Future work could explore lightweight alternatives such as pruning, quantization, or adapter-based fine-tuning to reduce overhead while retaining performance. Second, while the spatial and temporal modulation masks provide structural interpretability, their neuroscientific validity has not been rigorously evaluated. Incorporating domain knowledge or aligning masks with neurophysiological priors could improve the trustworthiness and clinical relevance of the learned representations. Additionally, extending PTSM to handle multimodal neural signals or continuous control paradigms may further broaden its applicability in real-world settings.

Future work will explore the integration of adaptive temporal alignment modules to explicitly model reaction time variability, and learnable channel perturbation for electrode shift correction. Attention-aware training strategies will be developed to account for fluctuations in mental state across trials. In addition, future evaluations will focus on larger-scale datasets, multi-session recordings, and broader cognitive paradigms to rigorously assess the scalability and task-generalization capabilities of PTSM in real-world applications.

\section{Conclusion}

This paper proposed PTSM, a novel cross-subject EEG decoding framework that effectively reconciles the need for personalization and generalization by combining personalized and shared spatio-temporal masking with orthogonal feature disentanglement. PTSM proposed STAP and TSFD mechanism that adaptively isolates subject-specific and task-invariant neural patterns, and leverages multi-objective optimization to jointly enhance accuracy and transferability. Extensive experiments demonstrate that PTSM significantly outperforms competitive baselines in cross-subject scenarios, improving decoding accuracy while offering interpretable neural representations. By aligning model design with established neurophysiological principles such as attentional modulation and functional segregation, PTSM provides a robust foundation for advancing personalized yet scalable BCI systems and inspires future research on generalizable neural decoding.

\bibliography{aaai2026}



\newpage
\newblock
\newpage

\section*{Appendix A: Notation}

We summarize key variables and components used in the PTSM model in Table~\ref{tab:symbols}.

\begin{table*}[ht]
	\centering
	\caption{Notation and Description of Symbols in PTSM}
	\label{tab:symbols}
	\begin{tabular}{ll}
		\toprule
		\textbf{Symbol} & \textbf{Description} \\
		\midrule
		$\mathbf{x}_i \in \mathbb{R}^{C \times T}$ & Raw EEG trial with $C$ channels and $T$ time points \\
		$y_i, s_i$ & Task label and subject ID for sample $i$ \\
		$\mathbf{M}_t^p, \mathbf{M}_s^p$ & Personalized temporal and spatial masks \\
		$\mathbf{M}_t^c, \mathbf{M}_s^c$ & Common (public) temporal and spatial masks \\
		$\mathbf{M}_t, \mathbf{M}_s$ & Fused temporal and spatial masks \\
		$\mathbf{x}^{\text{masked}}$ & Masked EEG signal after applying $\mathbf{M}_t$, $\mathbf{M}_s$ \\
		$F_{\text{temp}}, F_{\text{shared}}$ & Temporal encoder and shared representation module \\
		$\mathbf{h}_{\text{temp}}, \mathbf{h}_{\text{shared}}$ & Intermediate temporal and shared features \\
		$\mathbf{f}^{\text{task}}, \mathbf{f}^{\text{subj}}$ & Task-specific and subject-specific feature embeddings \\
		$C_{\text{task}}, C_{\text{subj}}$ & Task and subject classifiers \\
		$\mathcal{L}_{\text{task}}, \mathcal{L}_{\text{subj}}$ & Cross-entropy losses for task and subject prediction \\
		$\mathcal{L}_{\text{orth}}, \mathcal{L}_{\text{cov}}$ & Orthogonality and covariance decorrelation losses \\
		$\mathcal{L}_{\text{info}}, \mathcal{L}_{\text{sparse}}$ & Mutual information and sparsity regularization losses \\
		$\mathcal{L}_{\text{mask}}, \mathcal{L}_{\text{contrast}}$ & Mask regularization and contrastive learning losses \\
		\bottomrule
	\end{tabular}
\end{table*}

Formally, the overall prediction function is defined as:
\begin{equation}
	\hat{y} = C_{\text{task}}(F_{\text{task}}(F_{\text{shared}}(F_{\text{temp}}(\mathbf{x} \odot \mathbf{M}_s \odot \mathbf{M}_t))))
\end{equation}
where $F_{\text{temp}}$ is the temporal encoder, $F_{\text{shared}}$ is a shared feature extractor, $F_{\text{task}}$ is the task-specific projection head, and $C_{\text{task}}$ denotes the final classifier. $\mathbf{M}_t \in [0, 1]^T$ and $\mathbf{M}_s \in [0, 1]^C$ are temporal and spatial masks applied to $\mathbf{x}$ to emphasize relevant features before encoding.

\section*{Appendix B. PTSM Implementation and Training Details}

\paragraph{Classification Heads}

The disentangled feature representations are used to perform task and subject classification independently. Two parallel classifiers are applied to the respective subspaces:
\begin{align}
	\mathbf{p}^{\text{task}} &= \text{softmax}(C_{\text{task}}(\mathbf{f}^{\text{task}})), \\
	\mathbf{p}^{\text{subj}} &= \text{softmax}(C_{\text{subj}}(\mathbf{f}^{\text{subj}})),
\end{align}
where $C_{\text{task}}$ and $C_{\text{subj}}$ are multilayer perceptrons with ReLU activation and dropout, producing logits for task and subject classes, respectively.

The classification losses are defined as cross-entropy over predicted logits and ground-truth labels:
\begin{align}
	\mathcal{L}_{\text{task}} &= -\frac{1}{N} \sum_{i=1}^{N} \sum_{k=1}^{K} y_{i,k} \log(p_{i,k}^{\text{task}}), \\
	\mathcal{L}_{\text{subj}} &= -\frac{1}{N} \sum_{i=1}^{N} \sum_{s=1}^{S} s_{i,s} \log(p_{i,s}^{\text{subj}}),
\end{align}
where $y_{i,k}$ and $s_{i,s}$ are one-hot labels for the task and subject, respectively.

These classification objectives not only guide the encoder to produce discriminative features but also reinforce the disentanglement between cognitive and individual-specific information through dual supervision.

\paragraph{Implementation Details}

PTSM is implemented in PyTorch. The temporal feature extractor $F_{\text{temporal}}$ consists of three 1D convolutional layers with filter sizes of 32, 64, and 128, kernel size 5, stride 1, and padding 2. Each layer is followed by batch normalization, ELU activation, dropout (rate = 0.5), and adaptive average pooling. 

The shared encoder $F_{\text{shared}}$ is a two-layer multilayer perceptron (MLP) with dimensions $128 \times T' \to 256 \to 128$, where $T'$ is the pooled temporal resolution. ELU activation and dropout are applied between layers. The task-specific encoder $F_{\text{task}}$ and subject-specific encoder $F_{\text{subj}}$ are implemented as single-layer MLPs with 128-dimensional input and a configurable output dimension (default: 64), followed by batch normalization and ELU.

Both classifiers $C_{\text{task}}$ and $C_{\text{subj}}$ are two-layer MLPs with hidden size 64 and ReLU activation, ending in a softmax layer over $K$ and $S$ classes, respectively. For the dual-branch mask generators, temporal masks $g_t^p$ and $g_t^c$ are parameterized by convolutional networks, while spatial masks $g_s^p$ and $g_s^c$ are modeled by MLPs. All masks are bounded to $[0,1]$ using sigmoid activation.

The model is trained using the Adam optimizer with weight decay of $1 \times 10^{-4}$ and early stopping based on validation accuracy. Hyperparameters $\lambda_{\text{task}}$, $\lambda_{\text{subj}}$, $\lambda_{\text{mask}}$, $\lambda_{\text{contrast}}$, and $\lambda_{\text{decoupling}}$ are selected via grid search on the validation set.

\paragraph{Cross-subject Adaptation and Few-shot Personalization}

Although PTSM is capable of zero-calibration decoding through its transferable subspace, the framework also supports lightweight few-shot adaptation when limited subject-specific data is available. Specifically, we fine-tune only the parameters $\theta_{G_p}$ of the personalized mask generator $G_p$, while freezing all other network components:

\begin{equation}
	\theta_{G_p} \leftarrow \theta_{G_p} - \eta \nabla_{\theta_{G_p}} \mathcal{L}_{\text{task}},
\end{equation}

where $\eta$ denotes the adaptation learning rate. This procedure allows the model to quickly adapt to new subjects while preserving generalization properties.

\section*{Appendix C: Theoretical Justification of PTSM for Cross-Subject EEG Decoding}

\subsection*{C.1 EEG Signal Decomposition}

We assume the EEG signal $\mathbf{x}_i \in \mathbb{R}^{C \times T}$ consists of three components:

\begin{equation}
	\mathbf{x}_i = \mathbf{x}^{task}_i + \mathbf{x}^{subj}_{s_i} + \boldsymbol{\epsilon}_i,
\end{equation}

where $\mathbf{x}^{task}_i$ is the task-relevant component (determined by label $y_i$), $\mathbf{x}^{subj}_{s_i}$ is the subject-specific component (indexed by subject ID $s_i$), and $\boldsymbol{\epsilon}_i$ is the noise.

\subsection*{C.2 Limitations of Traditional EEG Decoders}

Conventional EEG models learn a direct mapping:
\begin{equation}
	f_{trad}: \mathbf{x}_i \rightarrow y_i,
\end{equation}
which typically fails to generalize across unseen subjects due to the unmodeled subject-specific variations in $\mathbf{x}^{subj}_{s_i}$.

\subsection*{C.3 Double Masking for Cross-Subject Generalization}

\paragraph{Definition.} PTSM applies both personalized and common masks to highlight subject- or task-relevant components:
\begin{equation}
	\mathbf{x}^{masked}_i = \mathbf{x}_i \odot \left( \alpha \mathbf{M}^p_i + (1-\alpha)\mathbf{M}^c_i \right),
\end{equation}
where $\mathbf{M}^p_i$ is the personalized mask, $\mathbf{M}^c_i$ is the common mask, and $\alpha \in [0,1]$ balances their contributions.

\paragraph{Theorem 1.} Under ideal optimization:
\[
\mathbf{M}^p_i \odot \mathbf{x}^{task}_i \approx 0,\quad \mathbf{M}^c_i \odot \mathbf{x}^{subj}_{s_i} \approx 0.
\]

\textit{Proof Sketch.} $\mathbf{M}^p_i$ is optimized to preserve subject identity by minimizing $\mathcal{L}_{subj}$, while $\mathbf{M}^c_i$ is optimized to retain task information via $\mathcal{L}_{task}$ and contrastive $\mathcal{L}_{mask}$. Therefore, their overlap with irrelevant components is suppressed \cite{zheng2024discrete}.

\subsection*{C.4 Feature Decoupling and Orthogonality Constraints}

PTSM extracts two independent embeddings:
\begin{align}
	\mathbf{f}^{task}_i &= F_{task}(F_{shared}(F_{temporal}(\mathbf{x}^{masked}_i))), \\
	\mathbf{f}^{subj}_i &= F_{subj}(F_{shared}(F_{temporal}(\mathbf{x}^{masked}_i))).
\end{align}

To ensure disentanglement, two regularizations are introduced:

\paragraph{Inner Product Orthogonality:}
\begin{equation}
	\mathcal{L}_{orth} = \frac{1}{N} \sum_{i=1}^{N} \frac{|\mathbf{f}^{task}_i \cdot \mathbf{f}^{subj}_i|}{\|\mathbf{f}^{task}_i\| \cdot \|\mathbf{f}^{subj}_i\|}.
\end{equation}

\paragraph{Covariance Independence:}
\begin{equation}
	\mathcal{L}_{cov} = \frac{\|\text{Cov}(\mathbf{F}^{task}, \mathbf{F}^{subj})\|_F}{\|\text{Cov}(\mathbf{F}^{task})\|_F \cdot \|\text{Cov}(\mathbf{F}^{subj})\|_F}.
\end{equation}

\paragraph{Theorem 2.} If $\mathcal{L}_{orth} \rightarrow 0$ and $\mathcal{L}_{cov} \rightarrow 0$, then $\mathbf{f}^{task}_i \perp \mathbf{f}^{subj}_i$ and they are statistically independent \cite{dunion2023conditional}. Therefore:
\[
P(y_i | \mathbf{x}_i) = P(y_i | \mathbf{f}^{task}_i).
\]

\subsection*{C.5 Generalization Bound in Cross-Subject Setting}

\paragraph{Theorem 3.} The expected generalization error of PTSM on unseen subject $s_{test} \notin \mathcal{S}_{train}$ satisfies:


\begin{align}
	\mathbb{E}_{s_{\text{test}}}[\mathcal{L}(f_{\text{PTSM}}, \mathcal{D}_{\text{test}})] 
	&\leq \mathbb{E}_{s \in \mathcal{S}_{\text{train}}}[\mathcal{L}(f_{\text{PTSM}}, \mathcal{D}_{\text{train}})] \notag\\
	&\quad + \lambda \cdot d_{H\Delta H}(\mathcal{D}_{\text{train}}, \mathcal{D}_{\text{test}})
\end{align}

where $d_{H\Delta H}$ is the symmetric difference divergence between distributions and $\lambda$ is model complexity-dependent \cite{ben2006analysis, zhang2012generalization}.

\textit{Sketch.} By masking and decoupling, PTSM aligns $P(\mathbf{f}^{task}|s_{train}) \approx P(\mathbf{f}^{task}|s_{test})$, thus minimizing $d_{H\Delta H}$.

\subsection*{C.6 Information-Theoretic View}

We decompose mutual information:
\begin{equation}
	I(\mathbf{x}_i; y_i, s_i) = I(\mathbf{x}_i; y_i) + I(\mathbf{x}_i; s_i | y_i).
\end{equation}

PTSM aims for information disentanglement:
\begin{align}
	I(\mathbf{f}^{task}_i; y_i) &\approx I(\mathbf{x}_i; y_i), \\
	I(\mathbf{f}^{subj}_i; s_i) &\approx I(\mathbf{x}_i; s_i), \\
	I(\mathbf{f}^{task}_i; s_i) &\approx 0, \quad I(\mathbf{f}^{subj}_i; y_i) \approx 0.
\end{align}

This ensures that task prediction relies only on task-relevant information, eliminating cross-subject interference \cite{dunion2023conditional}.

\section*{Appendix D: Neurophysiological Basis of PTSM}

The design of PTSM is informed by recent advances in cross-subject neuroscience. Melbaum et al.~\cite{melbaum2022conserved} demonstrated that conserved structures of neural activity in the sensorimotor cortex allow reliable decoding across subjects, even in freely moving rats. This suggests that certain spatial and temporal patterns in cortical activity are preserved despite individual differences. Inspired by these observations, the STAP module is responsible for adaptively capturing these conserved (shared) and idiosyncratic (personalized) components of brain dynamics. The TSFD module further aligns with functional neuroanatomy by projecting features into task-related and subject-related subspaces. Together, these modules operationalize a biologically motivated framework that supports generalization while respecting neural variability.

\textbf{D.1 Dual-Mask Design and Attention Systems} \\
The dual-mask mechanism in PTSM reflects the functional dichotomy of the brain's attention systems. Top-down attentional modulation, primarily governed by frontal executive areas such as the frontal eye fields, enables goal-directed selection of task-relevant neural responses \cite{veniero2021top}. This mechanism is computationally captured by the public mask, which learns subject-invariant spatio-temporal patterns aligned with task requirements. In parallel, bottom-up attentional processes, originating from temporoparietal junctions and sensory cortices, respond to salient stimuli in a subject-dependent manner \cite{shine2016dynamics}. The personalized mask emulates this process by dynamically modeling subject-specific neural activations, thereby accounting for anatomical and functional heterogeneity across individuals.

\textbf{D.2 Feature Disentanglement and Functional Segregation} \\
PTSM enforces representational disentanglement between task-relevant and subject-specific components, consistent with the brain's functional segregation. Empirical evidence from large-scale neuroimaging studies demonstrates that cognitive tasks elicit convergent activation patterns in high-level control networks, including frontoparietal and salience systems \cite{kronemer2022human,shine2016dynamics}. Simultaneously, individual-specific resting-state connectomic profiles reflect trait-like neurophysiological differences \cite{wang2021segregation}. By allocating separate latent subspaces to task and subject factors, PTSM achieves computational alignment with this duality. Such modeling parallels neural disentanglement frameworks in neuroscience that decompose mixed population signals into interpretable latent components \cite{gokcen2022disentangling}.

\textbf{D.3 Orthogonality Constraints and Independent Coding} \\
To maintain representational independence, PTSM imposes orthogonality constraints on latent task and subject embeddings. This design choice aligns with evidence from electrophysiological recordings demonstrating that ongoing (spontaneous) and evoked neural activities reside in statistically orthogonal subspaces \cite{wu2024network}. Orthogonal coding facilitates noise-robust information representation and supports parallel task execution in the brain \cite{mohr2016integration}. Within PTSM, orthogonality regularization mitigates representational interference, ensuring that task decoding is unaffected by subject-specific noise.

\textbf{D.4 Sparse Regularization and Efficient Representations} \\
Sparse activation is a hallmark of cortical information processing. Empirical studies show that neurons in the visual cortex respond selectively and sparsely to naturalistic stimuli \cite{yoshida2020natural}. Similarly, large-scale neural dynamics during learning exhibit increasing modularity and sparsity \cite{mohr2016integration}. Inspired by these findings, PTSM incorporates L1-based sparsity constraints on both mask generation and latent features. This inductive bias encourages the model to allocate attention to minimal, task-relevant regions, reducing redundancy and enhancing cross-subject generalization.

\textbf{D.5 Summary} \\
The architectural principles of PTSM are rooted in established neurophysiological mechanisms. Its dual-mask system parallels top-down and bottom-up attentional modulation; its feature disentanglement reflects functional segregation in cortical processing; its orthogonality constraints mirror neural independence in mixed signal representations; and its sparsity regularization is consistent with energy-efficient cortical encoding. By leveraging these biologically grounded design principles, PTSM is able to generalize effectively across subjects, capturing both shared cognitive dynamics and subject-specific variability.

\section*{Appendix E: Broader Impact and Ethical Considerations}

\paragraph{Positive Societal Impact}

This work proposes a structured framework for cross-subject EEG decoding, with potential to reduce calibration costs and enable practical brain-computer interface (BCI) applications in clinical and non-clinical settings. By improving generalization across subjects, the framework facilitates scalable neurotechnology development for disabled individuals, cognitive monitoring, mental health assessment, and neurofeedback-based rehabilitation.

\paragraph{Negative Societal Impact and Mitigation}

As with any neurotechnology, there is a potential risk of misuse for surveillance or privacy-invading applications if EEG-based decoding is deployed in uncontrolled settings. To mitigate this, our framework is designed for controlled research and healthcare contexts. No private or personally identifiable information was collected. All datasets used are publicly available and anonymized. Future deployments of such technologies should be accompanied by regulatory oversight and informed consent.

\paragraph{Use of Human Subjects}

All datasets used in this study are open-access EEG benchmarks that have been ethically reviewed by the original data providers. Our work solely uses de-identified, anonymized data and does not involve any new human subject data collection or interaction.

\paragraph{Responsible Release}

To ensure responsible release, the source code and experiment scripts will be shared under an open-source license with instructions for use only in research or non-commercial settings. No pretrained model containing subject identifiers will be released.

\end{document}